\documentclass[conference]{IEEEtran}
\IEEEoverridecommandlockouts
\usepackage{times}
\usepackage{epsfig}
\usepackage{graphicx}
\usepackage{amsmath}
\usepackage{amssymb}
\usepackage{indentfirst}
\usepackage{array}
\usepackage{multirow}
\usepackage{float}
\usepackage{hhline}
\usepackage{subcaption}
\usepackage{url}
\usepackage{fancyhdr}
\def\BibTeX{{\rm B\kern-.05em{\sc i\kern-.025em b}\kern-.08em
    T\kern-.1667em\lower.7ex\hbox{E}\kern-.125emX}}

\fancypagestyle{firstpage}{%
  \lhead{Accepted in 2019 International Joint Conference on Neural Networks (IJCNN)}
  \lfoot{\small{\copyright2019 IEEE. Personal use of this material is permitted. Permission from IEEE must be obtained for all other users, including reprinting/republishing this material for advertising or promotional purposes, creating new collective works for resale or redistribution to servers or lists, or reuse of any copyrighted components of this work in other works.}}
}

\begin{document}

\title{A Performance Evaluation of Convolutional Neural Networks for Face Anti Spoofing}

\author{\IEEEauthorblockN{Chaitanya Nagpal\IEEEauthorrefmark{1} and
Shiv Ram Dubey\IEEEauthorrefmark{2}}
\IEEEauthorblockA{Computer Vision Group,
Indian Institute of Information Technology, Sri City, India\\
\IEEEauthorrefmark{1}chaitanya.n14@iiits.in,
\IEEEauthorrefmark{2}srdubey@iiits.in}}

\maketitle
\thispagestyle{firstpage}

\begin{abstract}
In the current era, biometric based access control is becoming more popular due to its simplicity and ease to use by the users. It reduces the manual work of identity recognition and facilitates the automatic processing. The face is one of the most important biometric visual information that can be easily captured without user cooperation in an uncontrolled environment. Precise detection of spoofed faces should be on the high priority to make face based identity recognition and access control robust against possible attacks. The recently evolved Convolutional Neural Network (CNN) based deep learning technique has proven as one of the excellent method to deal with the visual information very effectively. The CNN learns the hierarchical features at intermediate layers automatically from the data. Several CNN based methods such as Inception and ResNet have shown outstanding performance for image classification problem. This paper does a performance evaluation of CNNs for face anti-spoofing. The Inception and ResNet CNN architectures are used in this study. The results are computed over benchmark MSU Mobile Face Spoofing Database. The experiments are done by considering the different aspects such as the depth of the model, random weight initialization vs weight transfer, fine tuning vs training from scratch and different learning rate. The favorable results are obtained using these CNN architectures for face anti-spoofing in different settings. 
\end{abstract}
\begin{IEEEkeywords}
Convolutional Neural Networks, Deep Learning, Face Anti-spoofing, Performance Evaluation, Inception, ResNet
\end{IEEEkeywords}

\section{Introduction}
In recent years, we have witnessed the growth and development of new and innovative methods for automatic authentication \cite{unar2014review}. With the growth of data and the increasing awareness about the sensitivity of personal information, people have started to treat their privacy more seriously. The development of more robust and user-friendly authentication and access control devices is on high priority by utilizing the visual information such as fingerprint, facial, iris, and many more as compared to the password and token based devices \cite{o2003comparing}. The major challenge in any automatic access control method is the protection against malicious attacks by intruders \cite{wayman2005introduction}. Specifically, in face based authentication, the major challenges are to deal with the following three attacks, (i) printed photos, (ii) replay videos, and (iii) 3D videos. Face anti-spoofing is the field of study that tackles the above mentioned challenges in a robust and efficient manner.

The face anti-spoofing is one of the fundamental problem of biometric and computer vision. In initial years, the hand-designed feature based approaches were more common and utilized the characteristics like texture-based features, motion-based features and depth-based features \cite{galbally2014biometric}. The texture-based analysis exploited the fact that real face contains different texture and illumination pattern as compared to a plastic or LCD surface used to accomplish the attack. Maatta et al. \cite{maatta} used a multi-scale local binary pattern (LBP) followed by a non-linear SVM to deal with such attacks. Chingovska et al. \cite{chingovska} also used the similar approach for the same problem. They extracted the LBP descriptors from a greyscale image and applied 3 classifiers on top of the LBP features to perform the classification. These methods are not efficient and require a lot of data pre-processing to be done. The motion based face anti-spoofing is also investigated by several researches by exploiting the fact that the most of face attacks happen with the use of stills and thus, lack the basic motion that can be used to differentiate a live subject from an image. Anjos et al. \cite{anjos} utilized the motion relation between foreground and background to differentiate between a live face and an attacked face. Pereira et al. \cite{pereira} used the LBP-TOP features containing space and time descriptors to encode the motion information along with the face texture. Kollreinder et al. \cite{kollreider} extracted the facial parts (e.g., left and right eyes, nose, left and right ears) by simplified optical flow and then modeled the liveliness of these parts through a short sequence of images. The noise in the face image is also treated as the important characteristics for face anti-spoofing with the fact that the noise level in attacked face is more due to the reconstruction process of any spoofing method. Zhang et al. \cite{zhang} utilized the multiple Difference of Gaussian (DoG) filters to remove the noise and low-frequency information. They used the high frequency information to generate the feature vector for SVM classifier to distinguish between genuine and fake faces. Wen et al. \cite{wen} considered the 4 types of surface deformations such as specular reflection, blurriness features, chromatic moment and color diversity to generate the feature vector and used SVM classifier to classify the feature vector into real vs spoofed. The above discussed methods had several drawbacks like the need to utilize hand designed features and the limited performance of these methods.

Recent trends in computer vision have shown a gradual shift towards Convolutional Neural Networks (CNN) due to its characteristics like automatic learning and higher accuracy \cite{jia2014caffe}. The CNN based approaches have been proven to be a very effective approach for different problems of visual information processing like object detection, semantic segmentation, image classification, biomedical analysis, image captioning, image coloring, biometric authentication, and many more \cite{gu2017recent}. In many scenarios, the performance of these methods even surpasses the human/expert level performance. ImageNet Large Scale Visual Recognition Challenge \cite{imagenet} has fostered the development of new and better CNN architectures over the years. The winning architectures like AlexNet \cite{alexnet} in 2012, VGGNet \cite{vggnet} and GoogleNet \cite{googlenet} in 2014, and ResNet \cite{resnet} in 2015 brought a number of improvements and innovations to the field of object recognition. The task of object detection has also witnessed a series of improvements over the last few years through the evolution of CNN architectures like Fast R-CNN \cite{fastrcnn} and Faster R-CNN \cite{fasterrcnn}. These approaches have made the object detection task not only faster than traditional methods but also improved the performance very drastically. The CNN architectures like Fully Convolutional Networks \cite{fullycnforsemanticsegmentation} and Mask R-CNN \cite{maskrcnn} have made the image segmentation much easier, intuitive and semantic. These approaches have gained very high improvement over its ancestral and hand-designed methodologies. The biomedical image processing area has also observed the immense improvement by using the CNN based methods in the problems like Colon Cancer Recognition \cite{coloncancernetwork}, \cite{basha2018rccnet} and Radiologist-Level Pneumonia Detection \cite{chexnet}, etc. The CNNs are also used for depth estimation from images \cite{repala2018dual}.

Some researchers have also explored CNNs for the biometric authentication and verification over the years as an alternative to traditional methods \cite{srivastava2019performance}. Different CNN architectures are proposed for different biometric traits such as fingerprint, face, iris, etc. \cite{menotti2015deep}, \cite{nogueira2016fingerprint}, \cite{taigman2014deepface}, \cite{zhang2016joint}, \cite{raja2015smartphone}. Facial authentication systems cover a number of problems as discussed earlier and various attempts have been made recently to solve these problems. Recently, CNN is also being applied for face anti-spoofing and liveliness detection. Gragnaniello et al. \cite{gragnaniello} utilized the domain-specific knowledge to deal with robustness problem in CNN architecture for biometric spoofing detection. Li et al. \cite{leili} fine tuned the CNN over face spoofing datasets and then extracted the features and applied the principle component analysis (PCA) to reduce the dimensionality and finally the SVM is employed to do the classification into real vs spoofed face. Atoum et al. \cite{yousef} utilized an ensemble of patch-based and depth-based CNN to perform the classification as well as liveliness detection in facial unlocking systems. All these methodologies proved that the CNNs can be used very effectively for the biometric anti-spoofing by automatically extracting the biometric features from training data.

Followings are the contributions of this paper:
\begin{itemize}
    \item Motivated by the success of CNNs in many visual information processing tasks in biometric and computer vision, this paper presents a performance evaluation of state-of-the-art CNN architectures such as Inception-v3, ResNet50 and ResNet152 for face anti-spoofing.
    \item  The experiments are conducted to cover the various aspects of using the CNN for face anti-spoofing such as the depth of architecture, fine tuning and training from scratch, pre-trained weight transfer and random weight initialization, and different learning rates.
    \item  This paper provides the best practices to utilize the CNN based approaches such as Inception-v3, ResNet50 and ResNet152 for face anti-spoofing.
\end{itemize}

The rest of the paper is divided into various sections. Section 2 discusses about the state-of-the-art CNN architectures compared in this study. Section 3 describes the experimental setup including the framework of face anti-spoofing using CNN, hyperparameter settings, the database characteristics and data preprocessing performed. Section 4 presents the experimental results with detailed analysis from different perspective. Sections 4 concludes the paper with constructive suggestions for future initiatives.\\

\section{CNN Architectures Used}
As discussed in the earlier section, the Convolutional Neural Networks (CNNs) are the new trends in computer vision. The CNNs have shown immense improvements in image and video based classification problems. In this study, we conduct a performance evaluation of state-of-the-art CNNs such as Inception-v3, ResNet50 and ResNet152 for face anti-spoofing. This section provides an overview of Inception and ResNet modules.

\begin{figure}[!t]
\begin{subfigure}{0.5\textwidth}
\centering
\includegraphics[width=\linewidth]{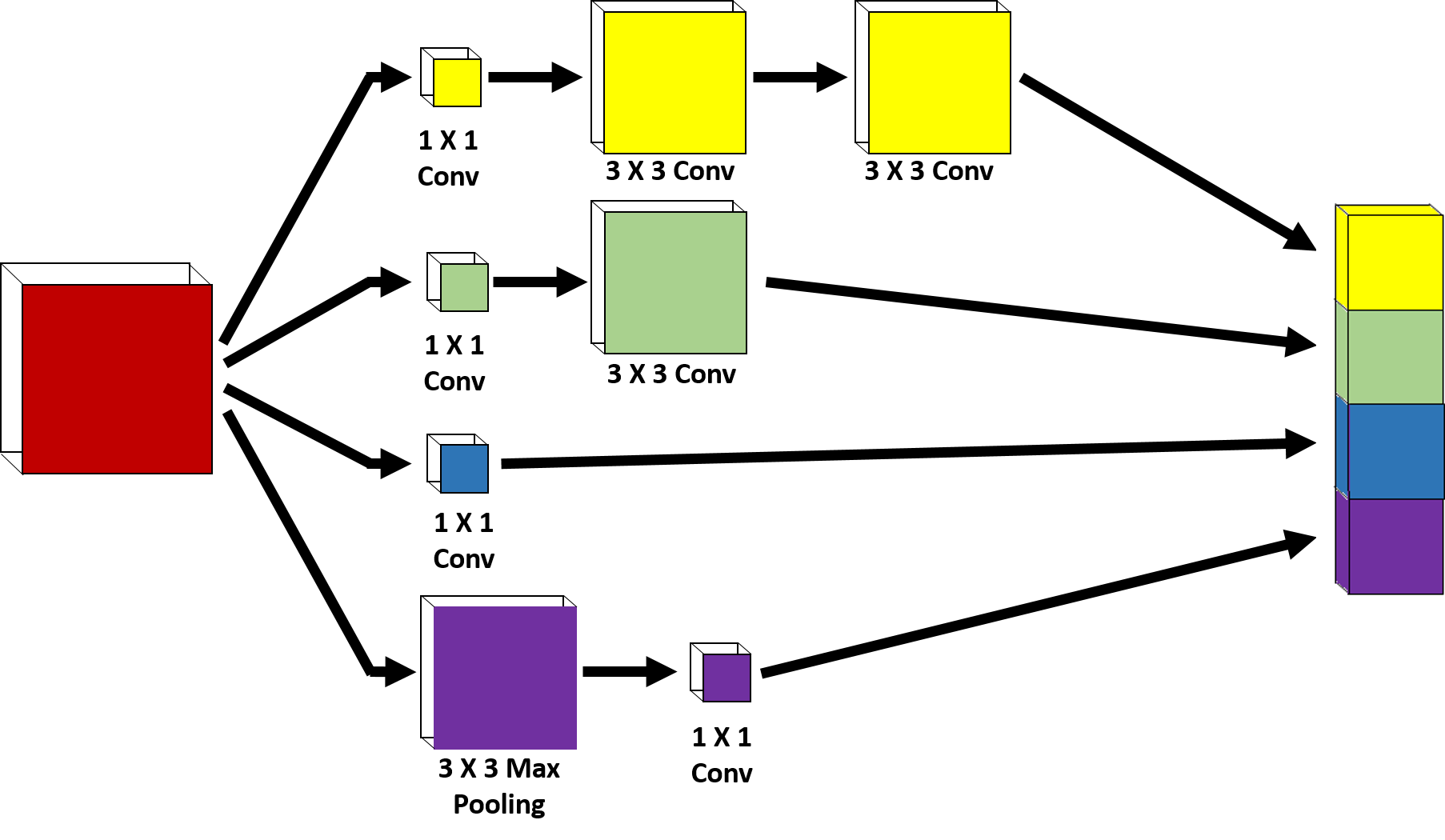}
\caption{Inception-v3 Module}
\end{subfigure}
\begin{subfigure}{0.5\textwidth}
\centering
\includegraphics[width=\linewidth]{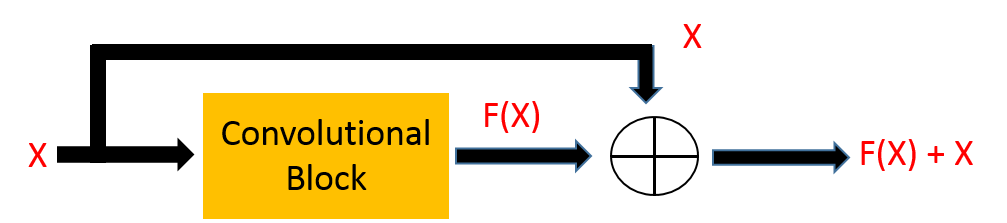}
\caption{Residual Module}
\end{subfigure}
\caption{The structure of Inception-v3 \cite{inceptionv3} and Residual \cite{resnet} modules. These modules are stacked to form the deep network of Inception-v3 \cite{inceptionv3} and ResNet \cite{resnet}, respectively.}
\label{fig:module}
\end{figure}

\subsection{Inception-v3 Module}
In 2014, Szegedy et al. of Google Inc. \cite{googlenet} proposed GoogLeNet which won the ImageNet Large-Scale Visual Recognition Challenge 2014 (ILSVRC14) \cite{imagenet} for classification and detection. The GoogLeNet is based on the inception module which basically combines the convolution outputs of varying filter sizes including $1\times1$, $3\times3$ and $5\times5$ with max pooling output. The original Inception module also uses the $1\times1$ bottleneck to reduce the complexity. Several inception modules are stacked over each other in GoogLeNet \cite{googlenet} to form a  22 layers deep network. The Inception module makes the GoogLeNet faster and efficient as compared to previous models like VggNet \cite{vggnet}, etc. Just after 1 year (i.e., in 2015), Szegedy et al. proposed an Inception-v3 module \cite{inceptionv3} which is basically the redesign version of the original Inception module \cite{googlenet}. The Inception-v3 module increases the computational efficiency drastically as compared to the original Inception module by factorization of the bigger convolutions into smaller convolutions. In Inception-v3 module, each $5\times5$ convolution is replaced by two $3\times3$ convolutions which reduces the number of operations while covering the same receptive field. The Inception-v3 module is shown in Fig. \ref{fig:module}(a). It computes four output volumes by applying (1) $1\times1$ convolution, (2) $1\times1$ convolution followed by $3\times3$ convolution, (3) $1\times1$ convolution followed by two $3\times3$ convolutions with different weights, and (4) $3\times3$ max pooling followed by $1\times1$ convolution, respectively for any input volume. Finally, all four output volumes are concatenated to form a single output volume. The dimension of output volume is same as the input volume by using the $1\times1$ bottleneck as depicted in Fig. \ref{fig:module}(a).

\subsection{Residual Module}
During the evolution of CNN architectures over the years from AlexNet (8 layers) \cite{alexnet} in 2012 to VggNet (16 or 19 layer) \cite{vggnet} and GoogLeNet (22 layers) \cite{googlenet} in 2014, it is observed that the deeper networks perform better due to the increased complexity. Following the same line, He et al. of Microsoft research \cite{resnet} conducted the experiment with 56-layer plain convolution architecture and found that the performance of 56-layer is worse than 20-layer. They analyzed that the deeper network is very hard to optimize and leads to the decreased performance. In order to overcome this optimization issue, they proposed to learn the residual instead of the plain transformation. They introduced the ResNet \cite{resnet} architecture which uses the residual block to pass more information towards the last layers. The residual unit basically facilitates to provide the crucial information to next unit which is actually lost in convolution step. 

The structure of residual unit is shown in Fig. \ref{fig:module}(b), here the convolutional block represent two convolution operation, $X$ is the input volume to residual unit, $F(X)$ is the output volume of convolutional block, and $F(X)+X$ is the output volume of residual block/unit. It can be perceived from Fig. \ref{fig:module}(b) that the residual unit learns $F(X)$ which is basically the residual of output from input. Several residual blocks are stacked in ResNet for deeper architecture. Based on the number of residual blocks, the depth of ResNet is different. The ResNet architecture was also the winner of ImageNet Large-Scale Visual Recognition Challenge 2015 (ILSVRC15) \cite{imagenet} classification task and achieved 3.57\% error which surpasses the human level performance. The ResNet also secured the $1^{st}$ positions for the ImageNet detection, ImageNet localization, COCO detection, and COCO segmentation tasks. In this study, ResNet50 and ResNet152 architectures with 50 and 152 layers, respectively are used for performance evaluation over face anti-spoofing.

\section{Experimental Setup}
This section describes the performance evaluation experimental setup in terms of the face anti-spoofing framework using CNN, hyperparameter settings, evaluation criteria and face spoofing database used.

\begin{figure*}[!t]
\begin{center}
\includegraphics[width=\linewidth]{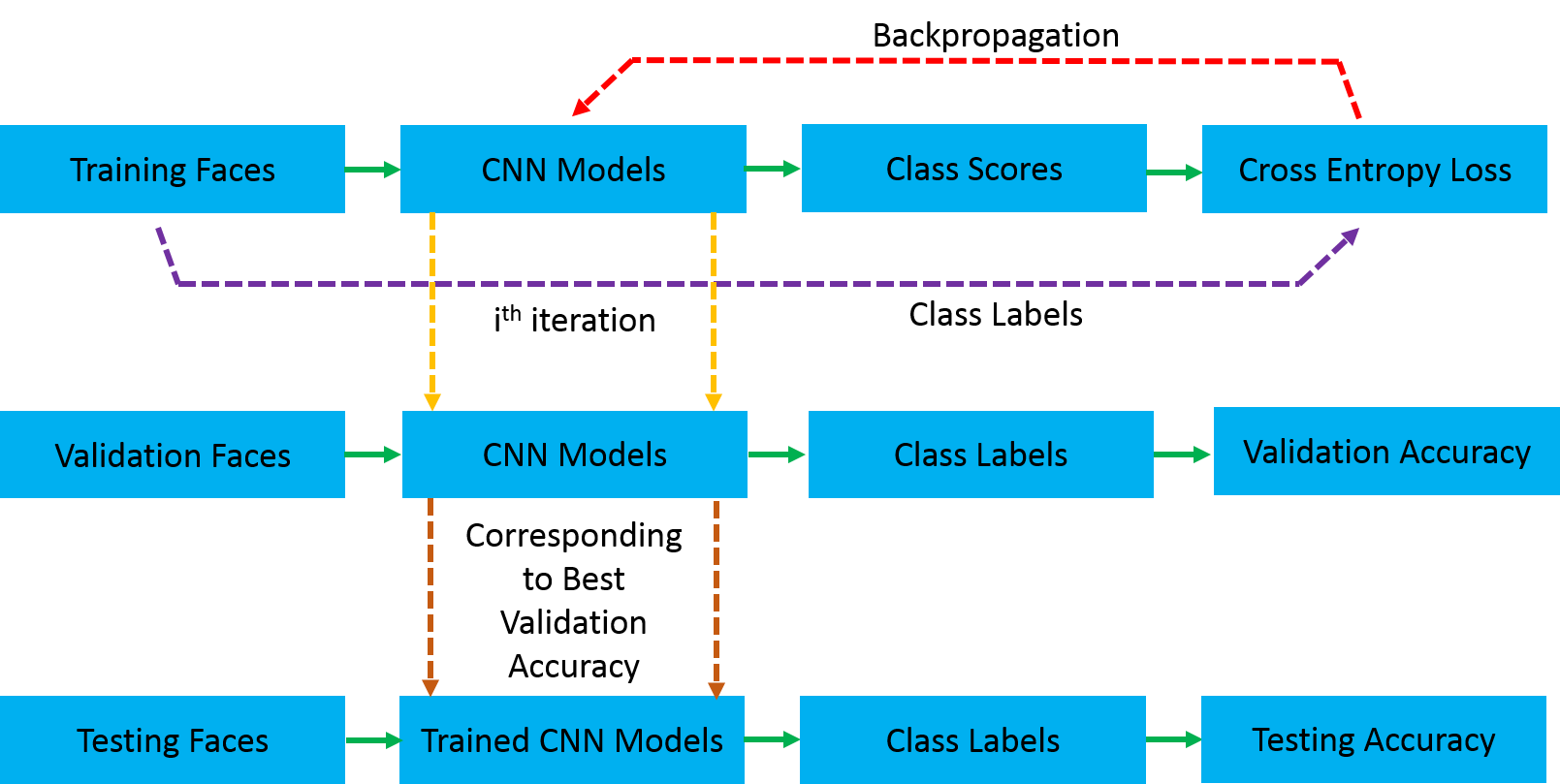}
\end{center}
\caption{Training, validation and testing framework for face anti-spoofing using CNN models. The Inception-v3, ResNet50 and ResNet152 models are used in this paper.}
\label{fig:fraamework}
\vspace{7mm}
\end{figure*}

\subsection{CNN based Face Anti-spoofing}
The face anti-spoofing is considered as the two-class classification problem in this paper. The two classes are real face class and spoofed face class. Fig. \ref{fig:fraamework} shows the training and testing framework for real and spoofed face classification using CNN model such as Inception-v3, ResNet50 and ResNet152. During training phase, the CNN model predicts the class score for training images, computes the categorical cross-entropy loss, and finally update the weights of network using gradient descent method by back-propagating the gradient w.r.t. loss function. In every epoch, the learned weights using training images are used to generate the class scores and classification accuracy over validation images. Once training is done, the learned weights corresponding to highest validation accuracy is used for testing. During the testing phase, the trained CNN model generates the class scores for input face image and predicts the class corresponding to the highest class score. 

The experiments are performed over a desktop computer system having an Intel Core i7-7700 CPU, 32 GB RAM (i.e., 2 $\times$ 16 GB RAM) and one 8 GB NVIDIA Zotac GeForce GTX 1080 GPU. The programs are written using the Keras open source neural network library in Python running on top of the TensorFlow deep learning framework. The Adam optimizer \cite{adam} has been proved to be a suitable by-default stochastic optimization technique in most of the problems of neural network. Thus, in our experiments also, the Adam optimizer is used to train all CNN models. The experiments are conducted with following different setups, (1) the weights are transferred from the pre-trained weights computed over ImageNet database, (2) the weights are initialized randomly, (3) only fully connected layer is trained and weights of other layers are freezed, (4) all layers are trained irrespective of the initialization, and (5) two learning rates (i.e., $10^{-3}$ and $10^{-5}$) are used without any learning rate annealing. In order to evaluate the performance of different models for different hyperparameter settings, the training, validation and testing accuracies are computed. The convergence time is also computed in terms of the minimum number of epochs needed to get the highest result.

\begin{table*}
\caption{The statistics of MFSD database \cite{wen} in terms of the number of samples in training, validation and testing sets along with the proportion of samples for different attacks.}
\newcolumntype{D}{>{\small\centering}p{0.1\linewidth}}
\newcolumntype{E}{>{\small\centering}p{0.06\linewidth}}
\newcolumntype{F}{>{\small\centering}p{0.06\linewidth}}
\begin{center}
\begin{tabular}{|D|E|E|E|E|E|E|E|F|E|F|}
\hline
& \multicolumn{7}{|c|}{Fake} & \multicolumn{3}{|c|}{Real}\tabularnewline
\hline
Device&Phone&Laptop&Phone&Laptop&Phone&Laptop&\multirow{2}{*}{Total}&Phone&Laptop&\multirow{2}{*}{Total}\tabularnewline
\cline{1-7}\cline{9-10}
Attack&Printed&Printed&Phone&Phone&Tablet&Tablet&&NA&NA&\tabularnewline
\hline\hline
Testing&410&376&507&516&546&516&2871&414&387&801\tabularnewline
\hline
Validation&411&350&469&486&447&485&2648&426&414&840\tabularnewline
\hline
Training&7201&6688&8862&9026&9461&9498&50736&7733&7812&15545\tabularnewline
\cline{2-11}
+Flipped&7201&6688&8862&9026&9461&9498&50736&7733&7812&15545\tabularnewline
\cline{2-11}
= Total & 14402&13376&17724&18052&18922&18996&101472&15466&15624&31090\tabularnewline 
\hline
\end{tabular}
\end{center}
\label{table:database}
\end{table*}

\begin{figure*}[!t]
\begin{center}
\includegraphics[width=\linewidth]{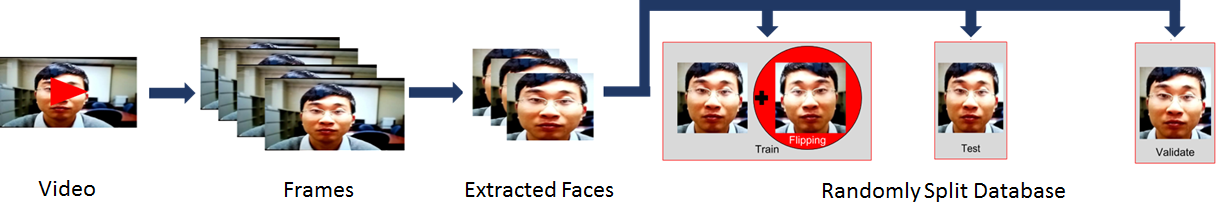}
\end{center}
\caption{MSU Mobile Face Spoofing Database (MFSD) \cite{wen} preparation including frames extraction from video, face localization in frames using Viola Jones Harr Cascade \cite{viola}, and sample split into training, validation and testing sets for the experiments in this study. Note that, only horizontal flipping is applied over training images for data augmentation.}
\label{fig:splitting}
\end{figure*}

\begin{figure*}[!t]
\begin{center}
\includegraphics[width=\linewidth]{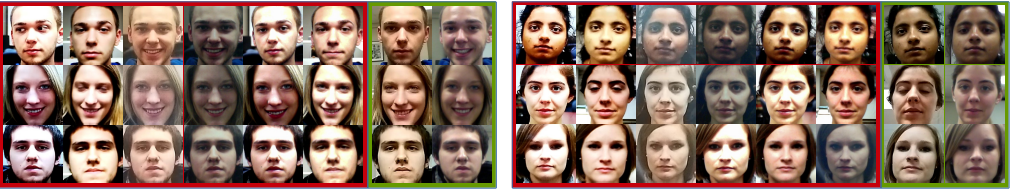}
\caption{The sample faces after cropping from MFSD database \cite{wen}. The faces in a row are corresponding to a particular subject. The images inside the Red and Green rectangular boxes contain the spoofed and real faces, respectively.}
\label{fig:samples}
\end{center}
\end{figure*}

\begin{center}
\begin{figure*}[!t]
\includegraphics[width=0.98\linewidth]{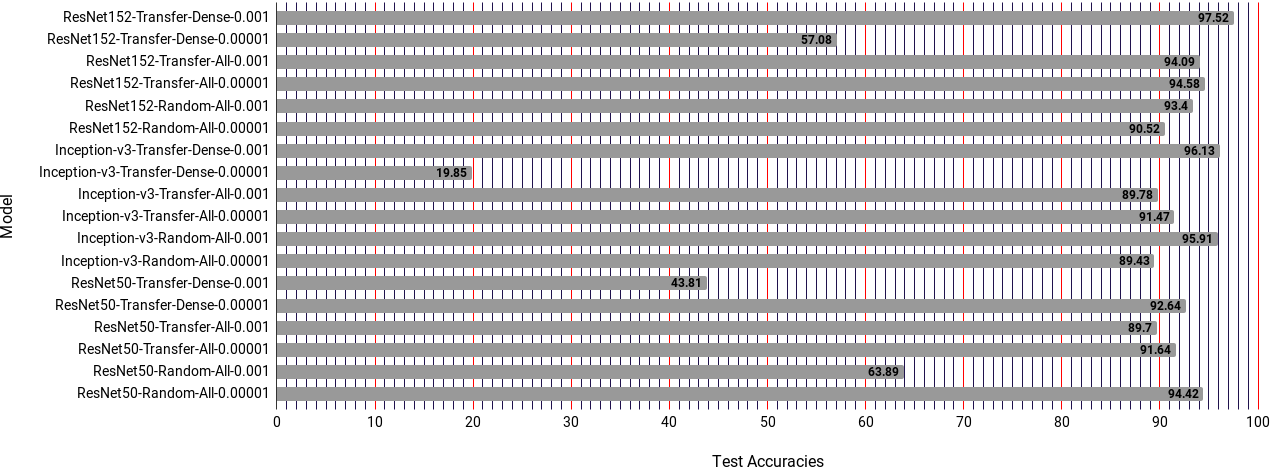}
\caption{Test accuracy corresponding to the trained weights of highest validation accuracy for different models explored in this study in different settings. Following is model name convention: ModelName-WeightInitializationType-TrainableLayers-LearningRate. Here, `Transfer' refers to the weight initialization by transferring from pre-trained ImageNet weights of that model, `Random' refers to random weight initialization, `Dense' corresponds to the training of dense layers only, and `All' corresponds to the training of all layers.}
\label{fig:test_acc}
\end{figure*}
\end{center}

\begin{figure*}[!t]
\begin{center}
\includegraphics[width=0.98\linewidth]{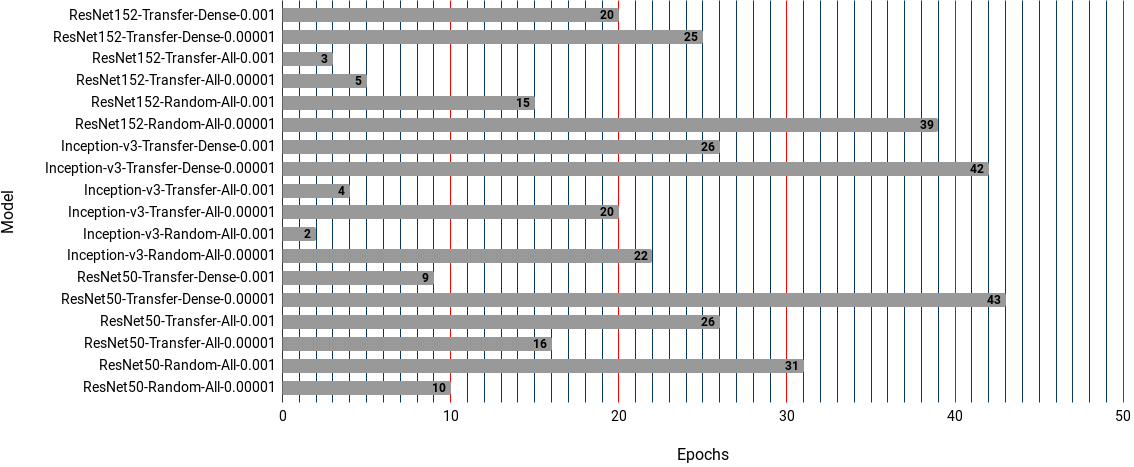}
\caption{The minimum number of epochs taken to reach the maximum validation accuracy. The naming convention is similar to Fig. \ref{fig:test_acc}.}
\label{fig:epochs}
\end{center}
\end{figure*}

\begin{center}
\begin{table*}[!t]
\centering
\caption{The training, validation and testing performance comparison among Inception-v3, ResNet50 and ResNet152 models in terms of the accuracy, convergence rate, and varying parameters like initial weights, number of trainable layers and learning rate. In this table, the `Epochs' is the number of epochs for highest validation accuracy.}
\newcolumntype{D}{>{\small\centering}p{0.09\linewidth}}
\newcolumntype{E}{>{\small\centering}p{0.14\linewidth}}
\newcolumntype{F}{>{\small\centering}p{0.11\linewidth}}
\newcolumntype{G}{>{\small\centering}p{0.07\linewidth}}
\begin{tabular}{|E|D|D|D|F|F|F|G|}
\hline
Base Model & Initial Weights & Trainable Layers & Learning Rate & Training Accuracy & Validation Accuracy & Testing Accuracy & Epochs  \tabularnewline
\hline
\hline
Resnet 152 & Imagenet & Dense & 0.001 & 92.63 & 99.59 & 97.52 & 20  \tabularnewline
\hline
Resnet 152 & Imagenet & Dense & 0.00001 & 94.06 & 55.96 & 57.08 & 25  \tabularnewline
\hline
Resnet 152 & Imagenet & All & 0.001  & 90.58 & 96.24 & 94.09 & 3 \tabularnewline
\hline
Resnet 152 & Imagenet & All & 0.00001 & 93.44 & 98.54 & 94.58 & 5 \tabularnewline
\hline
Resnet 152 & Random & All & 0.001 & 90.86 & 97.94 & 93.40 & 15 \tabularnewline
\hline
Resnet 152 & Random & All & 0.00001& 93.38 & 94.18 & 90.52 & 39 \tabularnewline
\hline
\hline
Inception-v3 & Imagenet & Dense & 0.001 & 94.63 & 96.47 & 96.13 & 26 \tabularnewline
\hline
Inception-v3 & Imagenet & Dense & 0.00001 & 91.88 & 20.56 & 19.85 & 42 \tabularnewline
\hline
Inception-v3 & Imagenet & All & 0.001 & 90.97 & 99.23 & 89.78 & 4 \tabularnewline
\hline
Inception-v3 & Imagenet & All & 0.00001 & 93.60 & 95.63 & 91.47 & 20 \tabularnewline
\hline
Inception-v3 & Random & All & 0.001 & 91.33 & 97.07 & 95.91 & 2 \tabularnewline
\hline
Inception-v3 & Random & All & 0.00001 & 93.95 & 98.57 & 89.43 & 22 \tabularnewline
\hline
\hline
ResNet 50 & Imagenet & Dense & 0.001 & 91.34 & 72.88  & 43.81 & 9 \tabularnewline
\hline
Resnet 50 & Imagenet & Dense & 0.00001 & 92.66 & 94.26 & 92.64 & 43 \tabularnewline
\hline
ResNet 50 & Imagenet & All & 0.001 & 91.26 & 98.71 & 89.70 & 26 \tabularnewline
\hline
ResNet 50 & Imagenet & All & 0.00001 & 96.67 & 98.22 & 91.64 & 16 \tabularnewline
\hline
ResNet 50 & Random & All & 0.001 & 93.96 & 97.62 & 63.89 & 31\tabularnewline
\hline
ResNet 50 & Random & All & 0.00001 & 92.80 & 97.88 & 94.42 & 10 \tabularnewline
\hline
\end{tabular}
\label{table:observations}
\end{table*}
\end{center}

\subsection{Database Used}
For the course of the performance evaluation in this paper, we used the benchmark MSU Mobile Face Spoofing Database (MFSD) \cite{wen}. It consists of 8 videos of 35 subjects. The video sets for each user consist of 2 real videos and 6 fake videos captured through various devices. For our experiments, first the videos are converted into frames, then the face in frames is localized by using the Viola Jones Harr Cascade \cite{viola}, and finally the extracted faces are randomly split into three sets including training, validation and testing. This procedure is depicted in Fig. \ref{fig:splitting}. The training set is flipped horizontally to apply the data augmentation. The sample faces extracted from MFSD database are displayed in Fig. \ref{fig:samples} before applying flipping. Each row corresponds to the faces of a particular subject. The spoofed and real faces are enclosed within the Red and Green rectangular boxes, respectively. 

The extracted face images are used to train, validate and test the Inception-v3, ResNet50 and ResNet152 models. A total of 73441 images are extracted from the videos of all subjects. These images are further distributed randomly into training, testing and validation sets including 66281, 3672 and 3488 images, respectively. The images in the training set are augmented by horizontal flipping, thus doubling the training dataset size to 132562. A complete statistics of the used MFSD database is presented in Table \ref{table:database} including the number of images of different attacks.

\section{Performance Evaluation and Observations}
In order to find the best practices for face anti-spoofing using CNN architectures such as Inception-v3, ResNet50 and ResNet152, we performed several experiments and the analyzed the results. We compared different CNN models in this section in terms of the accuracies, rate of convergence and other factors such as weight transfer, random weight initialization, fine tuning, training from scratch and different learning rate. 

\subsection{Test Accuracy Comparison}
The different models trained on the same database with varying parameters have shown drastic variances in performance. The Fig. \ref{fig:test_acc} shows the comparison among the accuracy of the Inception-v3, ResNet50 and ResNet152 models obtained over the test set corresponding to the highest validation accuracy. Highest test accuracy observed over the MFSD database is 97.52\% for the ResNet152 model trained through fine tuning of dense layers using ImageNet challenge ResNet152 weights at a learning rate of $10^{-3}$. For the same ResNet152, the test accuracy decreases on decreasing the learning rate in case of fine tuning of dense layers. However, the performance of ResNet152 increases after decreasing the learning rate when all the layers are trained. Comparing the weight initialization methods for ResNet152, it is evident that the test accuracy increases for weight transfer as compared to the random weight initialization while keeping all the other settings same. On the other hand, for Inception-v3 model, the results slightly vary. The highest test accuracy noted for Inception-v3 is 96.13\% which is achieved when the model is fine tuned on Imagenet challenge weights at a learning rate of $10^{-3}$. Even in the case of Inception-v3, it can be observed that by decreasing the learning rate for the same parameters causes an increase in the accuracy in general. However, when the Inception-v3 is fully trained with Imagenet weights, a higher accuracy is achieved at lower learning rate (i.e., $10^{-5}$). Comparing the weight initialization methods for Inception-v3, we observed that the random initialization works better for higher learning rate while transfer learning works better for lower learning rate. ResNet50 achieves highest accuracy of 94.42\% when trained from random weights at a learning rate of $10^{-5}$. In general it can be observed that ResNet50 performs better when trained with a lower learning rate.

\subsection{Convergence Rate Comparison}
The training of different CNN models exhibit the varying rate of convergence as shown in the Fig. \ref{fig:epochs}. It is affected by several factors like the model type, model complexity, model size, number of trainable layers, training method, etc. In general, the transfer learning is proved to be faster than random weight initialization based training for the same model. The Inception-v3 and ResNet50 models experience the gain in training time for transfer learning at lower learning rate of $10^{-5}$. It is evident from the  Fig. \ref{fig:epochs} that the ResNet152 model takes the most amount of time when initialized with random weights and trained with a learning rate of $10^{-5}$. One important observation of ResNet50 model is that when trained at a learning rate of $10^{-5}$, the model converges faster as compared to the learning rate of $10^{-3}$ while the training is dome through transfer learning with initial weights transferred from Imagenet challenge. On an average the Inception-v3, ResNet50 and ResNet152 models take about 24.1, 17.8, 18.17 epochs, respectively to converge.

\subsection{Training, Validation and Testing Results Comparison}
The training, validation and testing accuracy of Inception-v3, ResNet50 and ResNet152 models is summarized in Table \ref{table:observations}. Comparing the validation accuracy, it can be observed that the highest validation accuracy registered is 99.59\% for ResNet152 through transfer learning with a learning rate of $10^{-3}$. The same setup also achieves the highest testing accuracy as discussed earlier. It is also observed that the least validation accuracy for both Inception-v3 and ResNet152 models is observed through transfer learning at a learning rate of $10^{-5}$. Whereas, under the same conditions, the ResNet50 performs very well with a validation accuracy of 94.26\%.

\section{Conclusion}
In this paper, a performance comparison is conducted for face anti-spoofing by using the CNN models. The recently discovered and state-of-the-art CNN architectures such as Inception-v3, ResNet50 and ResNet152 are used in this study. The experiments are performed over MSU Mobile Face Spoofing Database (MFSD). The MFSD database is partitioned into training, validation and testing sets. The results are computed against the epoch number corresponding to the highest validation accuracy achieved. The performance comparison is done w.r.t. different conditions such as depth of ResNet model, weight initialization methods, number of trainable layers and learning rate. The ResNet152 model is the best suited one for face anti-spoofing task when only dense layers are trained with weight initialization through ImageNet weight transfer and learning rate of $10^{-3}$. It is also observed that the lower learning rate is better for ResNet152, whereas higher learning rate is better for ResNet50. The Inception-v3 gives an acceptable trade-off between accuracy and rate of convergence. It is also revealed from the results that the transfer learning over all layers leads to the faster rate of convergence for ResNet152 and Inception-v3 models, whereas the same setting is against the ResNet50 model. Based on the observations of this study of face anti-spoofing using CNN models, it is suggested to utilize the deeper models at lower learning rates with transfer learning for last fully connected layers. It is recommended to use Inception-v3 architecture with the similar setting of above mentioned ResNet152 such as transfer learning for dense layers at lower learning rate with limited computational resources.

{\small
\bibliographystyle{IEEEtran}
\bibliography{Reference}
}

\end{document}